\documentclass[12pt]{article}

\usepackage[margin=1in]{geometry}

\usepackage{setspace}

\usepackage{graphicx}

\usepackage{enumitem}

\usepackage{microtype}
\usepackage{subfigure}
\usepackage{booktabs}
\usepackage{comment}
\usepackage{wrapfig}
\usepackage{arydshln}
\usepackage{enumitem}
\usepackage{multirow}
\usepackage{url}
\usepackage{natbib}
\usepackage{authblk}

\RequirePackage[colorlinks,citecolor=blue,linkcolor=blue,urlcolor=blue,pagebackref]{hyperref}

\usepackage{amsmath}
\usepackage{amssymb}
\usepackage{mathtools}
\usepackage{amsfonts}
\usepackage{bm}
\usepackage{bbm}

\usepackage{algorithm}
\usepackage{algorithmic}

\def\eqref#1{equation~\ref{#1}}
\def\1{\bm{1}}

\DeclareMathAlphabet{\mathsfit}{\encodingdefault}{\sfdefault}{m}{sl}
\SetMathAlphabet{\mathsfit}{bold}{\encodingdefault}{\sfdefault}{bx}{n}

\def\calA{{\mathcal{A}}}

\def\calR{{\mathcal{R}}}

\def\calX{{\mathcal{X}}}
\def\calY{{\mathcal{Y}}}

\def\bbR{{\mathbb{R}}}

\DeclareMathOperator*{\argmin}{arg\,min}

\newcommand{\p}[1]{\left(#1\right)}

\newcommand{\bigp}[1]{\big(#1\big)}

\newcommand{\Bigp}[1]{\Big(#1\Big)}

\newcommand{\Biggp}[1]{\Bigg(#1\Bigg)}

\newcommand{\abs}[1]{\left|#1\right|}

\newcommand{\Exp}[1]{\mathbb{E}\left[#1\right]}
\newcommand{\bigExp}[1]{\mathbb{E}\big[#1\big]}

\newcommand{\BigExp}[1]{\mathbb{E}\Big[#1\Big]}

\usepackage{amsthm}

\theoremstyle{plain}

\usepackage[textsize=tiny]{todonotes}
\usepackage{multirow}
\usepackage{wrapfig}
\usepackage{subfigure}

\renewcommand{\eqref}[1]{(\ref{#1})}

\usepackage{xcolor}
\newcount\Comments  % 0 suppresses notes to selves in text
\newcommand{\kibitz}[2]{\ifnum\Comments=1\textcolor{#1}{#2}\fi}

\usepackage[capitalize,noabbrev]{cleveref}

\allowdisplaybreaks

\title{A Unified Theory for Causal Inference: Direct Debiased Machine Learning via Bregman-Riesz Regression}

\author{Masahiro Kato\thanks{Email: \texttt{mkato-csecon@g.ecc.u-tokyo.ac.jp}}$\,$}

\affil{The University of Tokyo}

\date{\today}
\begin{document}

\maketitle

\begin{abstract}
    This note introduces a unified theory for causal inference that integrates Riesz regression, covariate balancing, density-ratio estimation (DRE), targeted maximum likelihood estimation (TMLE), and the matching estimator in average treatment effect (ATE) estimation. In ATE estimation, the balancing weights and the regression functions of the outcome play important roles, where the balancing weights are referred to as the Riesz representer, bias-correction term, and clever covariates, depending on the context. Riesz regression, covariate balancing, DRE, and the matching estimator are methods for estimating the balancing weights, where Riesz regression is essentially equivalent to DRE in the ATE context, the matching estimator is a special case of DRE, and DRE is in a dual relationship with covariate balancing. TMLE is a method for constructing regression function estimators such that the leading bias term becomes zero. Nearest Neighbor Matching is equivalent to Least Squares Density Ratio Estimation and Riesz Regression.
\end{abstract}

\section{Introduction}
This note is written to convey and summarize the main ideas of \citet{Kato2025directbias,Kato2025directdebiased,Kato2025nearestneighbor}. These works propose the direct debiased machine learning (DDML) framework, which unifies existing treatment effect estimation methods such as Riesz regression, covariate balancing, density-ratio estimation (DRE), targeted maximum likelihood estimation (TMLE), and the matching estimator. For simplicity, we consider the standard setting of average treatment effect (ATE) estimation \citep{Imbens2015causalinference}. Note that the arguments in this note can also be applied to other settings, such as estimation of the ATE on the treated (ATT). For details, see \citet{Kato2025directdebiased}. Throughout this study, we explain how the existing methods mentioned above can be unified from the viewpoint of targeted Neyman estimation via generalized Riesz regression, also called Bregman-Riesz regression.

Specifically, these existing methods aim to estimate the nuisance parameters that minimize the estimation error between the oracle Neyman orthogonal score and an estimated Neyman orthogonal score. From this point of view, we can interpret Riesz regression, DRE, and the matching estimator as methods for estimating the Riesz representer, also called the bias-correction term or the clever covariates. Covariate balancing is in a dual relationship with these methods. TMLE is a method for regression function estimation to minimize the estimation error.

This generalization not only provides an integrated view of various methods proposed in different fields but also offers a practical guideline for choosing an ATE estimation algorithm. For example, for specific choices of basis functions and loss functions, we can automatically attain the covariate balancing property.

\section{Setup}
Let $(X, D, Y)$ be a triple of $k$-dimensional covariates $X \in \calX \subseteq \bbR^k$, treatment indicator $D \in \{1, 0\}$, and outcome $Y \in \calY \subseteq \bbR$, where $\calX$ and $\calY$ are the corresponding spaces, and $D = 1$ denotes treated while $D = 0$ denotes control. Following the Neyman-Rubin framework, let $Y(1) \in \calY$ and $Y(0) \in \calY$ be the potential outcomes for treated and control units. Let us define the ATE as
\[\tau_0 \coloneqq \bigExp{Y(1) - Y(0)}.\]

We observe $n$ units with $\{(X_i, D_i, Y_i)\}_{i=1}^n$, where $(X_i, D_i, Y_i)$ is an i.i.d. copy of the predefined triple $(X, D, Y)$. Our goal is to estimate $\tau_0$ using the observations.

\paragraph{Notations and assumptions} For simplicity, we assume that the covariate distributions of the treated and control groups have probability densities. We denote the probability density of covariates in the treated group by $p(x \mid D = 1)$, and that of the control group by $p(x \mid D = 0)$. We also denote the marginal probability density by $p(x)$ and the joint probability density of $(X, D)$ by $p(x, d)$. We introduce the propensity score, the probability of receiving treatment, by
\[e_0(X) \coloneqq \frac{p(x, 1)}{p(x)}.\]
For $d \in \{1, 0\}$, we denote the expected outcome of $Y(d)$ conditional on $X$ by $\mu_0(d, X) = \bigExp{Y(d) \mid X}$.

To identify the ATE, we assume unconfoundedness, positivity, and boundedness of the random variables; that is, $Y(1)$ and $Y(0)$ are independent of $D$ given $X$, there exists a universal constant $\epsilon \in (0, 1/2)$ such that $\epsilon < e_0(X) < 1 - \epsilon$, and $X$, $Y(1)$, and $Y(0)$ are bounded.

\section{Riesz Representer and ATE Estimators}
\label{sec:rieszate}
In ATE estimation, the following quantity plays an important role:
\[\alpha_0(D, X) \coloneqq \frac{D}{e_0(X)} - \frac{1 - D}{1 - e_0(X)}.\]
This term is referred to by various names in different methods. In the classical semiparametric inference literature, it is called the bias-correction term \citep{Schuler2024introductionmodern}. In TMLE, it is called the clever covariates \citep{vanderLaan2006targetedmaximum}. In the debiased machine learning (DML) literature, it is called the Riesz representer \citep{Chernozhukov2022automaticdebiased}. It may also be referred to as balancing weights \citep{Imai2013estimatingtreatment,Hainmueller2012entropybalancing}, inverse propensity score \citep{Horvitz1952generalization}, or density ratio \citep{Sugiyama2012densityratio}.

This term has several uses. First, if we know the function $\alpha_0$, we can construct an inverse probability weighting (IPW) estimator as
\[\widehat{\tau}^{\text{IPW}} \coloneqq \frac{1}{n}\sum_{i=1}^n \alpha_0(D_i, X_i)Y_i.\]
This is known as one of the simplest unbiased estimators for the ATE $\tau_0$. Another usage is bias correction. Given an estimate $\widehat{\mu}(d, X)$ of $\mu_0(d, X)$, we can construct a naive plug-in estimator as
\[\widehat{\tau}^{\text{PI}} \coloneqq \frac{1}{n}\sum_{i=1}^n \bigp{\widehat{\mu}(1, X_i) - \widehat{\mu}(0, X_i)}.\]
Such a naive estimator often includes bias caused by the estimation of $\mu_0$ that does not vanish at the $\sqrt{n}$ rate. Therefore, to obtain an estimator of $\tau_0$ with $\sqrt{n}$ convergence, we debias the estimator as
\[\widehat{\tau}^{\text{OS}} \coloneqq \frac{1}{n}\sum_{i=1}^n \Bigp{\alpha_0(D_i, X_i)\bigp{Y_i - \widehat{\mu}(D_i, X_i)} + \widehat{\mu}(1, X_i) - \widehat{\mu}(0, X_i)}.\]
This estimator is called the one-step estimator. There exists another direction of bias correction, called TMLE. In TMLE, we update the initial regression function estimates $\widehat{\mu}$ as
\[\widetilde{\mu}(d, X_i) = \widehat{\mu}(d, X_i) + \frac{\sum_{i=1}^n \alpha_0(D_i, X_i)\bigp{Y_i - \widehat{\mu}(D_i, X_i)}}{\sum_{i=1}^n \alpha_0(D_i, X_i)^2} \alpha_0(d, X_i).\]
Then, we redefine the ATE estimator as
\[\widehat{\tau}^{\text{TMLE}} \coloneqq \frac{1}{n}\sum_{i=1}^n \bigp{\widetilde{\mu}(1, X_i) - \widetilde{\mu}(0, X_i)}.\]

Thus, the term $\alpha_0$ plays an important role. When $\alpha_0$ is unknown, its estimation becomes a core task in causal inference, along with the usually unknown regression function $\mu_0$. We can view Riesz regression, DRE, covariate balancing, and the matching estimator as methods for estimating $\alpha_0$ with different loss functions. In addition, TMLE has a close relationship with these estimation methods from the targeted Neyman estimation perspective, explained below.

\section{Targeted Neyman Estimation}
Following the debiased machine learning literature, we refer to $\alpha_0$ as the Riesz representer. We also focus on the Neyman orthogonal score, defined as
\[\psi(X, D, Y; \mu, \alpha, \tau) \coloneqq \alpha(D, X)\bigp{Y - \mu(D, X)} + \mu(1, X) - \mu(0, X) - \tau.\]
From efficiency theory, we know that an estimator $\widehat{\tau}^{\text{oracle}}$ is efficient if it satisfies
\[\frac{1}{n}\sum_{i=1}^n \psi(X, D, Y; \mu_0, \alpha_0, \widehat{\tau}^{\text{oracle}}) = 0.\]
However, if we plug in estimators of $\mu_0$ and $\alpha_0$, we might incur bias caused by estimation. Even though the Neyman orthogonal score has the property that such bias can be asymptotically removed at a fast rate, it is desirable to construct estimators $\widehat{\mu}$ and $\widehat{\alpha}$ that behave well in finite samples.

Based on this motivation, \citet{Kato2025directdebiased} proposes the targeted Neyman estimation procedure, which aims to estimate $\mu_0$, $\alpha_0$, and $\tau_0$ such that the following Neyman error becomes zero:
\[L(\mu, \alpha, \tau) \coloneqq \frac{1}{n}\sum_{i=1}^n \psi(X, D, Y; \mu, \alpha, \tau).\]

This Neyman error can be decomposed as follows:
\begin{align*}
    L(\mu, \alpha, \tau) &= \frac{1}{n}\sum_{i=1}^n \psi(X_i, D_i, Y_i; \mu, \alpha, \tau)\\
    &= \frac{1}{n}\sum_{i=1}^n \Bigp{\bigp{\alpha_0(D_i, X_i) - \alpha(D_i, X_i)} \bigp{Y_i - \mu_0(D_i, X_i)} - \alpha(D_i, X_i)\bigp{Y_i - \mu(D_i, X_i)}\\
    &\quad - \bigp{\mu(1, X_i) - \mu(0, X_i)} + \tau}.
\end{align*}
Therefore, in expectation, we have
\begin{align*}
    &\Exp{L(\mu, \alpha, \tau)}\\
    &= \Exp{\frac{1}{n}\sum_{i=1}^n \Bigp{\alpha_0(D_i, X_i) - \alpha(D_i, X_i)} \Bigp{Y_i - \mu_0(D_i, X_i)}}\\
    &\quad + \Exp{\frac{1}{n}\sum_{i=1}^n \Bigp{\tau - \bigp{\mu(1, X_i) - \mu(0, X_i)}}}.
\end{align*}
Thus, the core error terms are the following two:
\begin{align}
\label{eq:first_error}
    &\frac{1}{n}\sum_{i=1}^n \Bigp{\alpha_0(D_i, X_i) - \alpha(D_i, X_i)} \Bigp{Y_i - \mu_0(D_i, X_i)},\\
\label{eq:second_error}
    &\frac{1}{n}\sum_{i=1}^n \Bigp{\tau - \bigp{\mu(1, X_i) - \mu(0, X_i)}}.
\end{align}

We can interpret Riesz regression, covariate balancing, and nearest neighbor matching as methods for minimizing the error in \eqref{eq:first_error} by estimating $\alpha_0$ well, while TMLE is a method that automatically sets \eqref{eq:second_error} to zero by estimating $\mu_0$ well. We further point out that these existing methods for estimating the Riesz representer $\alpha_0$ can be generalized using the Bregman divergence and the duality of loss functions.

\section{Bregman-Riesz Regression}
This section reviews Bregman-Riesz regression, proposed in \citet{Kato2025directbias} and \citet{Kato2025directdebiased}, which is also called generalized Riesz regression. Bregman-divergence regression generalizes Riesz regression in \citet{Chernozhukov2024automaticdebiased} from the viewpoint of DRE via Bregman divergence minimization. As pointed out in \citet{Kato2025directdebiased}, we can derive covariate balancing methods as the dual of the Bregman divergence loss by extending the results in \citet{BrunsSmith2025augmentedbalancing} and \citet{Zhao2019covariatebalancing}. Note that this duality depends on the choice of models: when using Riesz regression, we need to use linear models for $\alpha_0$; when using Kullback-Leibler (KL) divergence, we need to use logistic models for $\alpha_0$.

\subsection{Bregman Divergence}
Our goal is to estimate $\alpha_0$ so that we can minimize
\[\frac{1}{n}\sum^n_{i=1}\Bigp{\alpha_0(D_i, X_i) - \alpha(D_i, X_i)} \Bigp{Y_i - \mu_0(D_i, X_i)}.\]
For simplicity, let us ignore the term $\Bigp{Y_i - \mu_0(D_i, X_i)}$. Then, our goal is merely to minimize the discrepancy between $\alpha_0(D_i, X_i)$ and $\alpha(D_i, X_i)$.

We first recap the Bregman divergence. Bregman divergence is defined via a differentiable and strictly convex function $g\colon \bbR \to \bbR$. Given $d\in\{1, 0\}$ and $x \in \calX$, let us define the following pointwise Bregman divergence between $\alpha_0(d, x)$ and $\alpha(d, x)$:
\[\text{BR}_g\bigp{\alpha_0(d, x)\mid \alpha(d, x)} \coloneqq g(\alpha_0(d, x)) - g(\alpha(d, x)) - \partial g(\alpha(d, x)) \bigp{\alpha_0(d, x) - \alpha(d, x)},\]
where $\partial g$ denotes the derivative of $g$. Taking the average over the distribution of $X$, we define the following average Bregman divergence:
\[\text{BR}_g\bigp{\alpha_0\mid \alpha} \coloneqq \BigExp{g(\alpha_0(D, X)) - g(\alpha(D, X)) - \partial g(\alpha(D, X)) \bigp{\alpha_0(D, X) - \alpha(D, X)}}.\]
Ideally, we want to estimate $\alpha_0$ by minimizing this average Bregman divergence, which is represented as
\[\alpha^* = \argmin_{\alpha\in \calA} \text{BR}^\dagger_g\bigp{\alpha_0\mid \alpha},\]
where $\calA$ is a hypothesis class of $\alpha_0$. If $\alpha_0 \in \calA$, then $\alpha^* = \alpha_0$ holds.

However, this formulation is infeasible because it includes the unknown $\alpha_0$. Surprisingly, by a simple computation, we can drop the unknown $\alpha_0$ and define an equivalent optimization problem as
\[\alpha^* = \argmin_{\alpha\in \calA} \text{B}_g\bigp{\alpha},\]
where
\[\text{B}_g\bigp{\alpha} \coloneqq \BigExp{ - g(\alpha(D, X)) + \partial g(\alpha(D, X)) \alpha(D, X) -  \Bigp{\partial g(\alpha(1, X)) - \partial g(\alpha(0, X))}}.\]

Finally, by replacing the expectation with sample approximations, we obtain the following feasible optimization problem for estimating the Riesz representer $\alpha_0$:
\[
\widehat{\alpha} \coloneqq \argmin_{\alpha \in \calA}\widehat{\text{B}}_g\bigp{\alpha}  + \lambda J(\alpha),
\]
where $J(\alpha)$ is some regularization function, and
\[
\widehat{\text{B}}_g(\alpha) \coloneqq \frac{1}{n}\sum^n_{i=1}\Bigp{ - g(\alpha(D_i, X_i)) + \partial g(\alpha(D_i, X_i)) \alpha(D_i, X_i) - \Bigp{\partial g(\alpha(1, X_i)) - \partial g(\alpha(0, X_i))}}.
\]

\subsection{Squared Loss}
We consider the following convex function:
\[g^{\text{LS}}(\alpha) = (\alpha - 1)^2.\]
Under this choice of $g$, the estimation problem is written as
\begin{align}
\label{eq:squaredloss}
    \widehat{\alpha} \coloneqq \argmin_{\alpha \in \calA}\widehat{\text{BR}}_{g^{\mathrm{LS}}}\bigp{\alpha} + \lambda J(\alpha),
\end{align}
where
\[\widehat{\text{BR}}_{g^{\mathrm{LS}}}\bigp{\alpha} \coloneqq \frac{1}{n}\sum^n_{i=1}\p{- 2\bigp{\alpha(1, X_i) + \alpha(0, X_i)} + \mathbbm{1}[D_i = 1]\alpha(1, X_i)^2 + \mathbbm{1}[D_i = 0]\alpha(0, X_i)^2}.\]

This estimation method corresponds to Riesz regression in debiased machine learning \citep{Chernozhukov2024automaticdebiased} and least-squares importance fitting (LSIF) in DRE \citep{Kanamori2009aleastsquares}. Moreover, if we define $\calA$ appropriately, we can yield nearest neighbor matching, as pointed out in \citet{Kato2025nearestneighbor}, which extends the argument in \citet{Lin2023estimationbased}.

\paragraph{Stable balancing weights} We can use various models for $\calA$. For example, we can use neural networks, though it is known to cause serious overfitting problems for this kind of DRE objective \citep{Rhodes2020,Kato2021nonnegativebregman}.

This study focuses on linear-in-parameter models for squared loss, defined as
\[\alpha(D, X) = \beta^\top\Phi(D, X),\]
where $\Phi \colon \{1, 0\} \times \calX \to \bbR^p$ is some basis function that maps $(D, X)$ to a $p$-dimensional feature space, and $\beta$ is a $p$-dimensional parameter. For such a choice of basis function, the dual of the problem \eqref{eq:squaredloss} can be written as
\begin{align*}
    &\min_{\alpha \in \bbR^n} \|\alpha\|^2_2\\
    &\text{s.t.}\ \ \sum^n_{i= 1}\alpha_i \Phi(D_i, X_i) - \p{\sum^n_{i=1}\Bigp{\Phi(1, X_i)  - \Phi(0, X_i)}} = \bm{0}_p,
\end{align*}
where $\bm{0}_p$ is the $p$-dimensional zero vector. 
Here, for simplicity, we let $\lambda = 0$ in this argument.

This optimization problem is the same as the one in stable balancing weights \citep{Zubizarreta2015stableweights}. This result is shown in \citet{BrunsSmith2025augmentedbalancing}, and \citet{Kato2025directdebiased} calls it automatic covariate balancing since we can attain the covariate balancing property without explicitly solving the covariate balancing problem.

\subsection{KL Divergence Loss}
\label{sec:empbalancing}
Next, we consider the following KL-divergence-motivated convex function:
\[g^{\mathrm{KL}}(\alpha) = (|\alpha| - 1)\log\p{|\alpha| - 1} - |\alpha|.\]
Then, we estimate $\alpha_0$ by minimizing the empirical objective:
\[
\widehat{\alpha} \coloneqq \argmin_{\alpha \in \calA}\widehat{\text{BR}}_{g^{\mathrm{E}}}\bigp{\alpha}  + \lambda J(\alpha),
\]
where
\[\widehat{\text{BR}}_{g^{\mathrm{E}}}\bigp{\alpha} \coloneqq \frac{1}{n}\sum^n_{i=1}\Bigp{\log\p{|\alpha(D_i, X_i)| - 1} + |\alpha(D_i, X_i)| - \log\p{\alpha(1, X_i) - 1} - \log\p{-\alpha(0, X_i) - 1}}.\]
For the derivation of this loss, see \citet{Kato2025directdebiased}. If we use $g(\alpha) = |\alpha|\log\abs{\alpha} - |\alpha|$ instead of $g^{\mathrm{KL}}$, the optimization problem aligns with LSIF in DRE \citep{Sugiyama2007covariateshift}. On the other hand, if we use $g^{\mathrm{KL}}$, the optimization problem aligns with the tailored loss in covariate balancing \citep{Zhao2019covariatebalancing}. Under this choice, we obtain the following duality result for entropy balancing weights \citep{Hainmueller2012entropybalancing}.

\paragraph{Entropy balancing weights} This study focuses on logistic models for KL-divergence loss, defined as
\[\alpha(D, X) = \mathbbm{1}[D = 1]r(1, Z) - \mathbbm{1}[D = 0]r(0, Z),\]
where $r(1, Z) = \frac{1}{e(X)}$, $r(0, Z) = \frac{1}{1 - e(X)}$, and
\[
e(X) \coloneqq \frac{1}{1 + \exp\bigp{-\beta^\top \Phi(Z)}}.
\]
Here, $\Phi \colon \calX \to \bbR^p$ is a basis function that does not include $D$, unlike the basis function for squared loss. Under this choice, we can write the optimization problem as
\begin{align*}
    \widehat{r} \coloneqq & \argmin_{r \in \calR} \frac{1}{n} \sum^n_{i=1} \Biggp{ \mathbbm{1}[D_i = 1]\p{ - \log \p{\frac{1}{r(1, X_i) - 1}} + r(1, X_i)}\\
    &\qquad\qquad\qquad\qquad + \mathbbm{1}[D_i = 0]\p{ - \log \p{\frac{1}{r(0, X_i) - 1}} + r(0, X_i)} },
\end{align*}
where $\calR$ is the set of functions $r$ defined above. This objective function is called the tailored loss in \citet{Zhao2019covariatebalancing}.

Then, as shown in \citet{Zhao2019covariatebalancing}, from the duality, it is known that this problem is equivalent to solving
\begin{align*}
    &\min_{w \in (1, \infty)^n} \sum^n_{i=1}(w_i - 1)\log (w_i - 1)\\
    &\text{s.t.} \quad \sum^n_{i=1}\Bigp{ \mathbbm{1}[D_i = 1]w_i \Phi(1, X_i) - \mathbbm{1}[D_i = 0]w_i \Phi(0, X_i) } = \bm{0}_p.
\end{align*}
This optimization problem aligns with that in entropy balancing \citep{Hainmueller2012entropybalancing}. Here, note that for estimated $\widehat{w}_i$, we can write $\widehat{\alpha}(D_i, X_i) = \mathbbm{1}[D_i = 1]\widehat{w}_i - \mathbbm{1}[D_i = 0]\widehat{w}_i$

\section{Implementation Suggestion}
In practice, one of our recommendations is the following procedure:
\begin{itemize}
    \item Estimate the regression function $\mu_0$ in some way.
    \item Model the Riesz representer using the logistic model $e(X) = \frac{1}{1 + \exp\bigp{-\beta^\top \Phi(Z)}}$.
    \item Estimate $r_0(D, X)$ as
    \begin{align*}
    \widehat{r} \coloneqq & \argmin_{r \in \calR}\frac{1}{n}\sum^n_{i=1}\Biggp{\mathbbm{1}[D_i = 1]\p{ - \log \p{\frac{1}{r(1, X_i) - 1}} + r(1, X_i)}\\
    &\qquad\qquad\qquad\qquad + \mathbbm{1}[D_i = 0]\p{ - \log \p{\frac{1}{r(0, X_i) - 1}} + r(0, X_i)}}(Y_i - \widehat{\mu}(D_i, X_i))^2,
    \end{align*}
    where $r(1, Z) = \frac{1}{e(X)}$, $r(0, Z) = \frac{1}{1 - e(X)}$. Here, we used weights $(Y_i - \widehat{\mu}(D_i, X_i))^2$, motivated by targeted Neyman estimation.
    \item Apply TMLE to $\widehat{\mu}$ and update it to $\widetilde{\mu}$, as in Section~\ref{sec:rieszate}.
\end{itemize}
That is, we recommend using entropy balancing to estimate the Riesz representer and applying TMLE to obtain the final ATE estimator. As shown above, both squared loss (Riesz regression) and KL divergence correspond to the same error minimization problem with different losses. On the other hand, KL divergence uses a basis function $\Phi(X)$ that depends only on $X$, while squared loss uses a basis function $\Phi(D, X)$ with an additional input. Although we can use logistic models for squared loss, we lose the covariate balancing property.

However, the combination of squared loss (Riesz regression) and linear models is also effective in important applications. As discussed in \citet{Kato2025nearestneighbor}, Riesz regression includes nearest neighbor matching as a special case. By changing the kernel (basis function), we can also derive various matching methods. Moreover, \citet{BrunsSmith2025augmentedbalancing} finds that under Riesz regression, we can write the Neyman orthogonal score as linear in $Y$, similar to standard OLS or Ridge regression.

\section{Conclusion}
This note presents a unified framework for causal inference by connecting Riesz regression, covariate balancing, density-ratio estimation, TMLE, and matching estimators under the lens of targeted Neyman estimation. Central to this framework is the estimation of the Riesz representer, which plays a crucial role in constructing efficient ATE estimators. We demonstrate that several existing methods can be interpreted as minimizing a common error term with different loss functions, and we propose a practical implementation that combines entropy balancing and TMLE. This unified view not only clarifies the relationships among these diverse methods but also provides guidance for applied researchers in choosing robust and better estimation strategies. For theoretical details and simulation studies, see \citet{Kato2025directbias,Kato2025directdebiased,Kato2025nearestneighbor}.

\bibliography{arXiv2.bbl}

\bibliographystyle{tmlr}

\onecolumn

\appendix

\end{document}